\documentclass[preprint,12pt,authoryear]{elsarticle}

\usepackage{lineno}
\modulolinenumbers[5]

\usepackage{amsmath,amssymb}
\usepackage{graphicx}
\usepackage{booktabs}
\usepackage{multirow}
\usepackage{color}
\usepackage[ruled,vlined]{algorithm2e} 
\usepackage{amsmath,amssymb}            
\usepackage{graphicx}                   
\usepackage{float}                     
\usepackage{geometry}

\usepackage{array}
\usepackage{caption}
\usepackage{booktabs} 
\usepackage{adjustbox} 
\renewcommand{\arraystretch}{1.2} 

\usepackage{subfigure}
\usepackage{natbib}

\usepackage{tabularx}
\usepackage{multirow}
\usepackage[section]{placeins}

\usepackage{booktabs}
\usepackage{tabularx}
\usepackage{array}
\usepackage{ragged2e}

\usepackage{hyperref}

\newcolumntype{Y}{>{\RaggedRight\arraybackslash}X}

\begin{document}

\begin{frontmatter}

\title{SharP: Soft and Hard Prompt-Guided Augmentation with LLM for Low Resource Fake News Detection\tnoteref{publication}}
\tnotetext[publication]{Author manuscript of the article published in \textit{Expert Systems with Applications} 309 (2026), 131178. The Version of Record is available at \url{https://doi.org/10.1016/j.eswa.2026.131178}. \textcopyright\ 2026. This manuscript version is made available under the \href{https://creativecommons.org/licenses/by-nc-nd/4.0/}{CC BY-NC-ND 4.0 license}.}

\author[label5,label1]{Ziyu Luo}
\ead{ziyu.luo@stu.cqu.edu.cn}

\author[label6]{Wei Li}
\ead{liwei00128@cqu.edu.cn}

\author[label1]{Haoping Huang}
\ead{haoping_huang@stu.cqu.edu.cn}

\author[label2]{Kecheng Liu}
\ead{k.liu@henley.ac.uk}

\author[label5,label1]{Min Gao\corref{cor1}}
\ead{gaomin@cqu.edu.cn}

\cortext[cor1]{Corresponding author.}

\address[label5]{Key Laboratory of Dependable Service Computing in Cyber Physical Society (Chongqing University), Ministry of Education, Chongqing 401331, China}
\address[label1]{School of Big Data and Software Engineering, Chongqing University, Chongqing 400044, China}

\address[label6]{Department of Plastic Surgery, Chongqing University Central Hospital, Chongqing Emergency Medical Center,  
Chongqing 400000, China}
\address[label2]{Informatics Research Centre, University of Reading, Reading, United Kingdom}

\begin{abstract}

Fake news detection under low resource conditions is challenged by the scarcity of labeled data and the difficulty of capturing subtle deceptive patterns. In this work, we focus on \emph{low resource} fake news detection, where only a small labeled set is available for training, while evaluation is performed under a fixed and unchanged test protocol. This setting is practically motivated by the high cost and ambiguity of expert fact-checking, as well as the rapid emergence of new topics and domains that are weakly covered by existing benchmarks, which together make large-scale, fully supervised training difficult to sustain.
Existing data augmentation methods, such as synonym substitution, generative adversarial networks, and large language model (LLM)-based generation, often produce fluent but overly generic content, lacking the task-specific relevance needed for accurate detection. To address this, we propose \textbf{SharP}, a boundary-aware text generation framework guided by \underline{s}oft and \underline{har}d \underline{p}rompts, which leverages LLMs as tools adapted to produce task-specific synthetic samples. SharP employs a dual prompt mechanism: hard prompts provide structured guidance and factual constraints, while soft prompts introduce learned semantic patterns from the data. These two types of prompts are combined to guide the generation process, enabling the model to produce samples that closely align with the characteristics of news. To further enhance detection performance, we incorporate a boundary-aware strategy that steers generation toward areas where the classifier is less confident, helping to clarify subtle distinctions between real and fake content. In addition, we adopt a dual objective optimization that balances semantic alignment with the source data and classifier feedback. This encourages the generation of samples that are both domain-consistent and helpful for refining decision boundaries, ultimately improving model robustness and generalization in low resource fake news detection.

\end{abstract}

\begin{keyword}
Fake News Detection \sep Large Language Models \sep Soft and Hard Prompting  \sep Data Augmentation \sep Low Resource Learning  

\end{keyword}

\end{frontmatter}

\section{Introduction}

The proliferation of fake news has emerged as a pressing global issue, driven by the unprecedented speed and reach of information dissemination enabled by social media platforms. Misinformation can severely distort public perception and trigger widespread consequences across domains such as politics, economics, and public health. Notable real-world examples include coordinated disinformation campaigns during elections and the circulation of false medical claims during the COVID-19 pandemic. Given its broad societal impact, fake news detection has become a focal research area across natural language processing, information retrieval, and computational journalism.

In recent years, a variety of techniques have been developed to tackle fake news detection, including textual content analysis, social propagation modeling, and multimodal fusion methods. However, the effectiveness of such models fundamentally relies on access to large-scale, high-quality labeled datasets. While existing benchmark datasets have facilitated significant progress, they remain limited in coverage, especially for emerging misinformation topics and early-stage detection tasks. Manual annotation of fake news is expensive, domain-specific, and subject to ambiguity, making the scalability of supervised learning approaches a persistent challenge. Consequently, enhancing model performance under low resource conditions has become a critical concern.

Data augmentation has emerged as a practical strategy to alleviate this bottleneck by expanding training sets with synthetic or transformed examples. Traditional text augmentation methods, such as synonym substitution, function word deletion, and back-translation, have shown effectiveness in improving model generalization~\citep{kapusta2024}. More recent efforts have introduced contrastive learning~\citep{hua2023}, multimodal synthesis, and GAN-based generation to further enrich the diversity of training data. With the advent of large-scale pre-trained language models, such as GPT and T5, researchers have begun to explore their potential for generating high-quality augmented samples in NLP tasks~\citep{choi2025}. These models offer strong semantic representation and generative capabilities, presenting a new frontier for data augmentation in fake news detection.

Nevertheless, existing LLM-based augmentation techniques often treat the generation process as a general-purpose paraphrasing task, lacking task-specific alignment and domain-level control. While these methods produce grammatically correct and fluent samples, they frequently fail to capture the deceptive stylistic patterns and subtle linguistic cues that distinguish fake news from real news. Moreover, most approaches do not consider the evolving weaknesses of the detection model during training. This results in a gap between the augmented data and the actual regions in the input space where the model is uncertain or error-prone.

Several recent studies have highlighted both the potential and the limitations of LLMs in this domain. \citet{chen2025} empirically demonstrated that LLMs are effective at identifying textual misinformation but struggle to model propagation structures and social context. \citet{raza2025} compared BERT-like encoders with autoregressive LLMs, finding that the former achieve better classification accuracy while the latter offer stronger robustness under noisy inputs. \citet{liu2025} proposed a few-shot LLM-based framework that integrates external evidence and reasoning chains to enhance detection under ambiguity. These works collectively point to the need for more targeted LLM utilization strategies in fake news detection.

This paper addresses a central question: \textit{How can LLMs be guided to generate augmented samples that are consistent with the domain, preserve the original labels, and effectively inform the classifier, thereby improving fake news detection in low resource settings?} We argue that effective augmentation must satisfy three key requirements: maintaining semantic fidelity to preserve the true labels, ensuring domain specificity to capture deceptive traits, and being sensitive to the model’s uncertain predictions in order to improve classification performance.

To this end, we propose \textbf{SharP}, a boundary-aware text generation framework guided by soft and hard prompts. SharP transforms LLMs into task-aligned generators through prompt engineering, active boundary sampling, and dual objective optimization. The core idea is to control both \textit{how} and \textit{where} samples are generated: not only by instructing the model to preserve stylistic and semantic constraints, but also by strategically augmenting high-uncertainty regions in the classifier’s decision space.

Specifically, we develop a dual prompt architecture that combines structured \textit{hard prompts}, which encode generation rules to enforce factual stance and stylistic fidelity, with \textit{soft prompt} that are continuously optimized to inject domain-aware inductive bias into the LLM’s attention layers. This enables fine-grained control over the generation process without altering the model's parameters. In addition, we introduce a boundary-sensitive dynamic augmentation strategy, which integrates an active learning loop to identify samples close to the classifier's decision boundary. These samples are often associated with the uncertainty of the model and provide the most informative training signals. By focusing augmentation on these regions, SharP not only increases data diversity but also enhances model discrimination where it is most needed. Finally, we propose a dual objective optimization mechanism that jointly enforces feature-level semantic alignment and classification performance. The first objective ensures that augmented samples remain close to the originals in the feature space, maintaining semantic fidelity. The second uses the classifier’s error feedback on generated samples to iteratively refine the soft prompts, guiding the augmentation process to improve task relevance.

In this work, we define \emph{low resource} fake news detection as a label-scarce training regime instantiated by restricting the amount of labeled training data. Concretely, we construct low resource settings by subsampling the labeled training set at multiple ratios, yielding progressively larger labeled pools while keeping the overall experimental setting consistent. All methods are trained under the same sampling ratios to enable a controlled comparison across different levels of supervision scarcity.

Taken together, SharP represents a principled integration of prompt-based control, uncertainty-driven sampling, and optimization-aware generation. By aligning augmentation with domain knowledge and classifier behavior, it substantially enhances model robustness and generalization in low resource fake news detection scenarios.

The contributions of this work are summarized as follows:
\begin{itemize}
    \item A prompt-guided LLM-based augmentation framework was developed, in which structured hard prompts were integrated with learnable soft prompts to enable domain-consistent and label-preserving generation across diverse news topics. This framework was implemented and empirically validated on the PHEME dataset.

   \item A boundary-aware sample selection strategy was proposed, leveraging classifier uncertainty to dynamically guide augmentation toward informative regions in the decision space. 

   \item A dual objective optimization scheme was designed to jointly promote semantic fidelity and discriminative enhancement, enabling end-to-end coordination between the augmentation module and the detection model. The contribution of this scheme was validated via hyperparameter sensitivity analysis, confirming its role in enhancing model performance.

\end{itemize}

The source code for SharP is available at \url{https://github.com/Fujizzz/SharP}.

\section{Related Work}

\subsection{Rumor Detection}

Rumor detection on social media has evolved rapidly as misinformation becomes more prevalent and time-sensitive. 
An early milestone in neural rumor detection was to model textual signals for timely identification; for example, \citet{gao2020} combined a character-level bidirectional language model with stacked LSTMs to capture sequential cues for early detection.
A subsequent milestone was the shift from plain-text modeling toward leveraging richer supervision signals, incorporating heterogeneous user-generated context and multimodal evidence.
For example, \citet{qian2021} proposed a cross-modal detection framework that integrates textual content, social comments, and image-level sentiment information to improve robustness under diverse scenarios.

More recently, the literature has increasingly emphasized robustness under limited supervision, where labeled rumors are scarce and newly emerging events deviate from past distributions. 
To alleviate label scarcity, \citet{sun2022} combined self-supervised learning with augmentation to improve representation stability in rumor detection. 
In parallel, feature fusion architectures have been advanced for platform-specific settings; \citet{wang2022} studied Chinese social media and combined textual, visual, and user-level features via ERNIE and VGG-19 to strengthen multimodal integration.

Overall, the field has progressed from sequence-based textual modeling to multimodal and context-aware learning, and further toward augmentation- and self-supervision-driven robustness. 
Despite substantial progress, rumor detection still faces a persistent bottleneck in low resource and evolving-event scenarios, where annotation is costly and ambiguous and models must learn from limited, shifting evidence—motivating the need for task-aligned augmentation under scarce supervision.

\subsection{Benchmark Datasets for Fake News and Rumor Detection}

To situate our study on low resource fake news detection, we briefly review representative benchmarks that are widely used to evaluate models under controlled label budgets. 
A major line of work focuses on social-platform rumor verification, where the input often comes with conversational or event structures, such as the breaking-news threads in PHEME~\citep{zubiaga2016} and the event-centric Twitter15/16 benchmarks~\citep{Ma2017Twitter15,Ma2017Twitter16}; complementary to English Twitter benchmarks, the Weibo rumor dataset provides a widely used Chinese microblog testbed, reflecting the linguistic and platform diversity of misinformation benchmarks~\citep{Ma2016Weibo}.
Another line targets claim/news-based fake news detection, including LIAR with fine-grained veracity labels for short fact-checked claims~\citep{Wang2017LIAR} and FakeNewsNet, which pairs news articles with associated social context from PolitiFact and GossipCop~\citep{Shu2020FakeNewsNet}. 
In addition, CoAID offers a domain-specific benchmark for COVID-19 healthcare misinformation~\citep{Cui2020CoAID}, enabling evaluation under rapidly evolving topics.
Importantly, these datasets differ in annotation granularity, instance structure (e.g., claim-, article-, or thread-level), and available auxiliary signals; therefore, low resource settings are typically instantiated by subsampling the labeled pool while keeping the evaluation protocol fixed.
This makes them well-suited for assessing whether augmentation methods can improve generalization when labeled supervision is scarce, which is the primary setting addressed in our work.

\newcommand{\groupline}[1]{%
\addlinespace[0.25em]
\multicolumn{5}{@{}l}{\textbf{#1}}\\
\addlinespace[0.10em]
}

\begin{table}[H]
\centering
\scriptsize
\setlength{\tabcolsep}{3.5pt}
\renewcommand{\arraystretch}{1.08}
\caption{Representative benchmark datasets for fake news and rumor detection}
\label{tab:dataset_overview}
\begin{tabularx}{\linewidth}{@{} p{0.16\linewidth} p{0.20\linewidth} p{0.10\linewidth} p{0.16\linewidth} X@{}}
\toprule
\textbf{Dataset} & \textbf{Source} & \textbf{Lang.} & \textbf{Scale} & \textbf{Labels / Structure} \\
\midrule

\groupline{Social rumor verification}
PHEME\par{\scriptsize\citep{zubiaga2016}} & 
Twitter events & EN & 330/4.8k & Conversation threads; rumor verification / rumor vs. non-rumor. \\

Twitter15/16\par{\scriptsize\citep{Ma2017Twitter15,Ma2017Twitter16}} & 
Twitter events & EN & 1490/818 & Event-level rumor detection with reply structure. \\

Weibo (rumor)\par{\scriptsize\citep{Ma2016Weibo}} & 
Sina Weibo & ZH & 2,560 (1280 fake + 1280 true) & Post-level rumor vs. non-rumor (common benchmark variants). \\

\addlinespace[0.2em]
\groupline{News/claim fake news detection}
LIAR\par{\scriptsize\citep{Wang2017LIAR}} & 
PolitiFact claims & EN & 12.8k & 6-way truthfulness; often binarized in detection. \\

FakeNewsNet\par{\scriptsize\citep{Shu2020FakeNewsNet}} & 
PolitiFact/GossipCop & EN & 1.1k/22.1k & Article-level fake/real with associated social context. \\

CoAID\par{\scriptsize\citep{Cui2020CoAID}} & 
COVID-19 content & EN & 1.9k (news articles) & Health misinformation benchmark (release-dependent labeling). \\

\bottomrule
\end{tabularx}
\end{table}

\subsection{Data Augmentation for Fake News Detection}

Data augmentation is a widely adopted strategy for improving text classification under limited labeled data, and it has been extensively explored in fake news detection. 
Early augmentation methods relied on syntactic or lexical perturbations. 
Back-Translation~\citep{sennrich2016} generates semantically preserved paraphrases through round-trip translation, and EDA~\citep{wei2019} further popularized lightweight operations such as synonym replacement, random insertion, swapping, and deletion. 
Subsequent work investigated interpolation-style augmentation at the representation level; MixText~\citep{feng2020} extends the Mixup principle to textual inputs by mixing latent features to synthesize new training instances. 
In addition, \citet{cheng2020} combined back-translation with adversarial training to promote diversity and robustness.

A major milestone in text augmentation is the emergence of large pre-trained language models, which shifts the paradigm from heuristic perturbations toward controllable, prompt-guided generation.
LM-CPPF~\citep{amirhossein2023} explored prompt-based few-shot paraphrasing with large models such as GPT-3 and OPT-175B, and \citet{himmet2024} proposed leveraging the fill-mask capability within Transformer-based models for augmentation. 
Despite these advances, existing augmentation methods often produce fluent yet task-agnostic samples and underutilize domain-specific deceptive cues, underscoring the need for controllable, label-preserving, and domain-aligned generation in low resource fake news detection.

\subsection{Soft Prompting in LLMs}

Parameter-efficient prompting has emerged as an appealing alternative to full-model fine-tuning for adapting large language models. 
Early work demonstrated that optimizing continuous prompt representations can effectively steer model behavior with a small number of trainable parameters. 
A milestone in this line is Prefix-Tuning~\citep{li2021}, which learns task-specific continuous prefix vectors to condition generation while keeping the backbone frozen.
Soon after, Prompt Tuning~\citep{lester2021} further demonstrated that optimizing continuous prompt embeddings alone can be competitive for downstream tasks, solidifying “trainable prompts” as a practical paradigm for parameter-efficient adaptation.
These techniques have been particularly attractive for few-shot and low resource settings due to their reduced training cost. 
Complementing prompt optimization, \citet{zhao2021} proposed contextual calibration to mitigate prediction biases in few-shot language models, improving accuracy and reducing sensitivity to prompt formats, demonstration selection, and demonstration order. 
More recently, \citet{zhu2024} proposed Soft Knowledgeable Prompt-Tuning (SKPT), integrating external knowledge to enhance short-text classification under limited-label regimes.

Overall, the literature has progressed from discrete prompt engineering to continuous, trainable prompting, and further toward incorporating structured priors such as external knowledge. 
Despite the effectiveness of prompting, directly applying it to fake news augmentation remains challenging because misinformation spans multiple domains and styles: hard prompts may lack adaptability, while soft prompts require task-aligned optimization signals to generate domain-consistent and label-preserving samples.

\begin{table}[t]
\centering
\caption{Representative LLM-related studies for fake news detection and their reported results}
\label{tab:llm_related}
\scriptsize
\setlength{\tabcolsep}{3pt}
\renewcommand{\arraystretch}{1.06}
\begin{tabularx}{\linewidth}{@{}p{0.20\linewidth} p{0.26\linewidth} Y@{}}
\toprule
Work & LLM role & Datasets / key results (as reported in the original papers) \\
\midrule
\citet{chen2025} &
Prompted LLM as detector &
On FND23, GPT-3.5-turbo with a task-specific prompt reports Acc=0.875 and Macro-$F_1$=0.528; the paper additionally benchmarks multiple datasets/prompting variants. \\

\citet{raza2025} &
LLMs vs.\ BERT-like (zero-/few-shot, fine-tune) &
The authors report $F_1$: on their GPT-labeled dataset vs.\ NELA-GT, RoBERTa achieves 0.897 (vs.\ 0.836), while Mistral-7B-v0.2 fine-tuned achieves 0.802 (vs.\ 0.737). \\

\citet{choi2025} &
LLM-based data augmentation &
Finds LLM-generated augmentation yields limited or inconsistent gains for COVID-19 misinformation stance detection across settings, cautioning that fluency does not guarantee task-useful diversity. \\

\midrule
\textbf{This work (SharP)} &
\textbf{Boundary-aware LLM-guided augmentation (hard + soft prompts)} &
\textbf{PHEME (5 events, 1\%--70\% labeled): Acc improves from 0.760 (1\%) to 0.938 (70\%); consistently outperforms strong baselines under low resource splits (see Table~\ref{tab:baseline_models}).} \\

\bottomrule
\end{tabularx}
\end{table}

\begin{figure}[h!]
    \centering
    \includegraphics[width=1.0\textwidth]{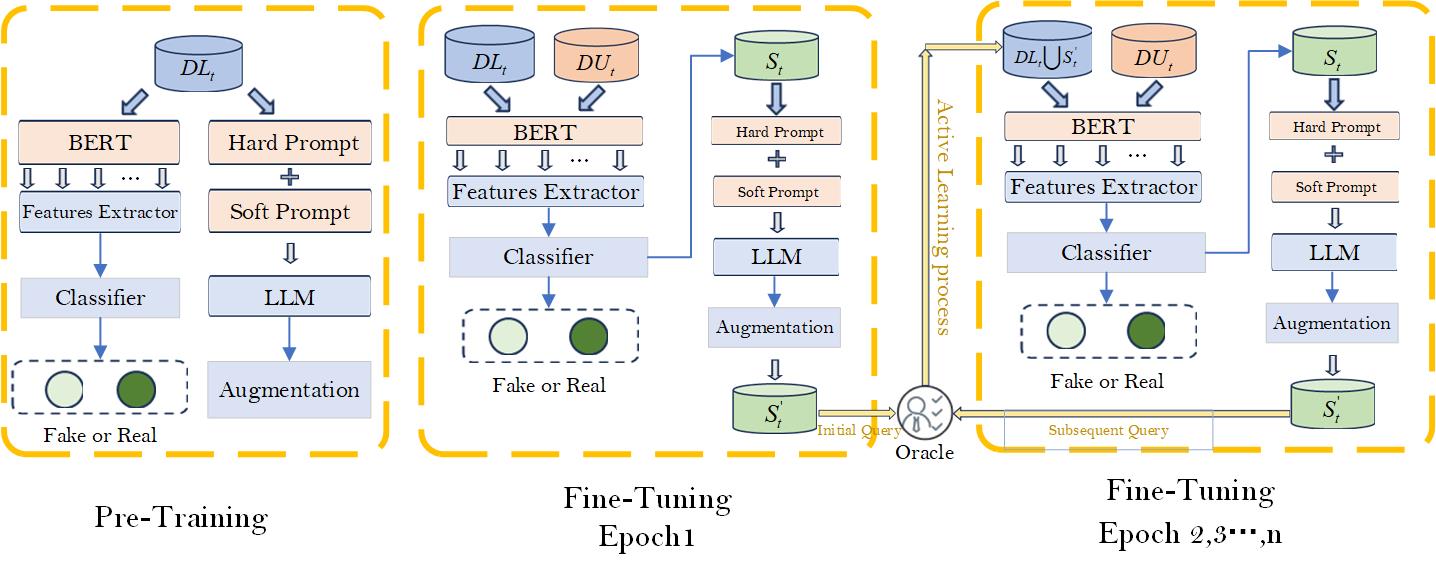}
    \caption{Model Architecture of SharP}
    \label{fig:model_architecture}
\end{figure}

\section{Method}

In this section, we first formally describe the problem, then introduce the key modules of the fake news detection model, and provide a detailed explanation of the data augmentation strategy and tuning methods. As shown in Figure~\ref{fig:model_architecture}, the proposed model consists of two core modules: the classifier module and the data augmentation module. The classifier module integrates a feature extractor and a classifier, responsible for identifying the authenticity of fake news. The data augmentation module, centered on an LLM, combines hard prompts and soft prompts to guide the model in generating augmented text, aiming to alleviate the problem of scarce training samples and improve the model's detection ability and generalization performance in low resource environments.

\begin{table*}[t]
\centering
\caption{Notation used in the Method section}
\label{tab:notation}
\footnotesize
\setlength{\tabcolsep}{5pt}
\renewcommand{\arraystretch}{1.05}
\begin{tabular}{@{}p{0.18\textwidth}p{0.78\textwidth}@{}}
\toprule
\textbf{Symbol} & \textbf{Description} \\
\midrule
$DL_t=(X_L,Y_L)$ & Labeled pool at round $t$, where $X_L$ and $Y_L$ denote labeled texts and their labels. \\
$DU_t=X_U$ & Unlabeled pool at round $t$, where $X_U$ denotes unlabeled texts. \\

$S_t$ & Queried unlabeled batch selected from $DU_t$ at round $t$. \\
$S_t'$ & Augmented labeled set generated from $S_t$ (inherits labels). \\
$b$ & Query batch size in each active learning round. \\
$T$ & Total number of active learning rounds. \\
$x_u^{(i)}$ / $T_{\text{in}}^{(i)}$ & Input text sequence of the $i$-th queried instance (we use $x_u^{(i)} \equiv T_{\text{in}}^{(i)}$ when convenient). \\

$x_u^{(i)\prime}$ / $T_{\text{out}}^{(i)}$ & Augmented (generated) text for the $i$-th queried instance (we use $x_u^{(i)\prime} \equiv T_{\text{out}}^{(i)}$ when convenient). \\

$\mathcal{H}(x)$ & Predictive entropy (uncertainty) of the classifier for input $x$. \\
$P(y\mid x)$ & Predicted posterior probability of $y\in\{0,1\}$ for input $x$. \\
$\tau$ & Entropy threshold for selecting boundary samples. \\
$\mathcal{B}_t$ & Boundary sample set at round $t$ selected by $\mathcal{H}(x)\ge \tau$. \\

$T_h$ & Hard prompt (discrete instruction tokens). \\
$T_s$ & Soft prompt vector (learnable continuous prefix). \\
$\oplus$ & Concatenation operator. \\
$\odot$ & Element-wise (Hadamard) product. \\

$T_{\text{concat}}$ & Input concatenated with hard prompt. \\
$E_{\text{combined}}$ & Combined embedding fed into the LLM (after adding $T_s$). \\
$H_{\text{out}}$ & Last-layer hidden representation of the LLM output. \\

$F(\cdot)$ & Feature extractor. \\
$D(\cdot)$ & Classifier. \\
$F_{\text{in}},F_{\text{out}}$ & Feature vectors of the input and generated representations. \\

$c,\;c^*$ & Classifier score and target score. \\
$\lambda_1,\lambda_2$ & Weights for combining the two objectives in prompt tuning. \\
$\lambda_{\text{reg}},\delta$ & Regularization weight and minimum-distance margin. \\
$\eta$ & Learning rate. \\
$N_1,N_2,N$ & Epochs for training $D/F$, training $T_s$, and fine-tuning $D$ per round. \\
\bottomrule
\end{tabular}
\end{table*}

\subsection{Problem Definition}

\begin{figure}[htbp]
    \centering
    \includegraphics[width=1\textwidth]{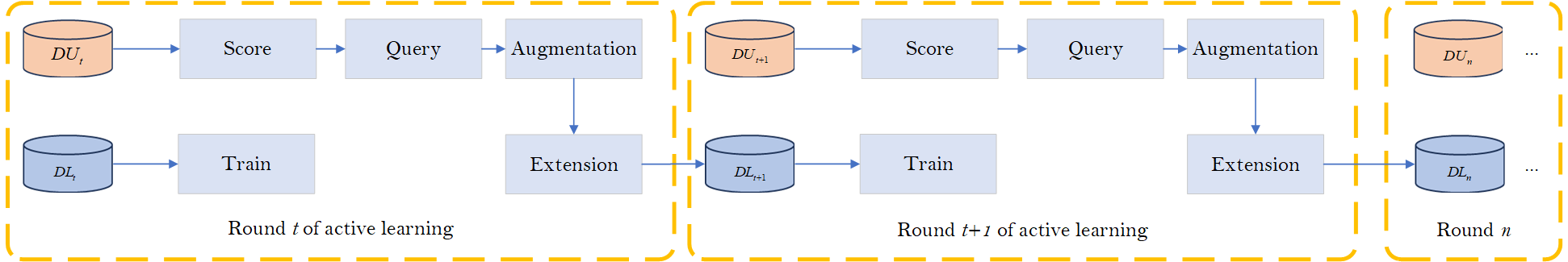}
    \caption{Method Flow}
    \label{fig:method_flow}
\end{figure}

Figure~\ref{fig:method_flow} summarizes the iterative learning process. We initialize the labeled sample pool as \(DL_0 = (X_L, Y_L)\) and the unlabeled
sample pool as \(DU_0 = X_U\). At the \(t\)-th round of active learning
(\(t = 0, \dots, T-1\)), we denote the current labeled and unlabeled pools by
\(DL_t\) and \(DU_t\), respectively. 
A mini-batch of unlabeled instances is queried for labeling, where \(b\) is the query batch size. After obtaining labels \(\{ y_u^{(i)} \}_{i=1}^{b}\) for the queried instances, we denote the labeled queried set as:
\begin{equation}
S_t = \left\{ (x_u^{(i)}, y_u^{(i)}) \right\}_{i=1}^{b}.
\end{equation}

The corresponding labels \(\{ y_u^{(i)} \}_{i=1}^{b}\) are obtained (e.g., from human annotators
or an auxiliary labeling model), and the classifier is subsequently retrained on
the enlarged labeled set. For each queried instance \(x_u^{(i)}\), the
augmentation module generates a synthetic counterpart \(x_u^{(i)\prime}\),
resulting in the augmented set:
\begin{equation}
S_t' = \left\{ (x_u^{(i)\prime}, y_u^{(i)}) \right\}_{i=1}^{b}
\end{equation}

where each synthetic sample inherits the label of its source instance, i.e.,
\((x_u^{(i)\prime}, y_u^{(i)})\).

Finally, the labeled and unlabeled pools are updated as follows:
\begin{equation}
\begin{aligned}
DL_{t+1} &= DL_t \cup S_t \cup S_t',\\
DU_{t+1} &= DU_t \setminus \{x \mid (x,y)\in S_t\}.
\end{aligned}
\end{equation}

\subsection{Active Learning Module}

We design a classifier model integrated with a feature extractor and combine it with an active learning strategy to select samples near the model's decision boundary, i.e., those for which the model exhibits the highest uncertainty in its predictions. These samples are highly valuable for model training and will be used for data augmentation in subsequent steps, further enhancing model performance.

To effectively extract emotional features, we construct emotion representations based on five categories of emotional lexicons: Emotion, Sentiment, Morality, Imageability, and Hyperbole. We use the NRC Emotion Lexicon~\citep{mohammad2013} to extract 8 emotional features, and the NRC Sentiment Lexicon and the Moral Foundations Dictionary~\citep{graham2009} to extract emotional and moral features (2 and 10, respectively). Additionally, we leverage the MRC Psycholinguistic Database~\citep{wilson1988} to extract 2 Imageability-related features and construct emotional features by analyzing approximately 350 hyperbole words.

Finally, these multi-dimensional emotional features are integrated into a unified emotional vector, representing the diverse emotional information in the news.

We design a dual feature extraction model that combines sentiment and textual features based on feature extraction. The model comprises two key components: the first component uses Convolutional Neural Networks to extract word-level latent representations of the text (\(v_{\text{word}}\)). Specifically, the input text sequence is encoded by BERT as \( S = \{w_1, w_2, \dots, w_n \} \), where \( w_i \) is the embedding of the \(i\)-th word. The CNN applies multi-scale convolutions to the word embeddings, using kernel sizes of 3, 4, and 5 to extract vertical features. The max pooling layer selects the maximum value from each feature map generated by the convolution operations, forming a pooled feature vector and producing the text's latent representation (\(v_{\text{word}}\)):

\begin{equation}
v_{\text{word}} = \text{MaxPool}(CNN(w_i)).
\end{equation}

The second component uses a Bidirectional Gated Recurrent Unit to capture sentiment flow in the text, producing a sentiment representation (\(v_{\text{affect}}\)). Here, \( e \) represents the sentiment vector:

\begin{equation}
v_{\text{affect}} = \text{Bi-GRU}(e).
\end{equation}

After obtaining the word-level representation (\(v_{\text{word}}\)) and sentiment features (\(v_{\text{affect}}\)), we combine them using element-wise multiplication to generate a unified text-sentiment representation (\(v_{\text{fuse}}\)). This step deeply fuses sentiment and text information, enhancing the model's ability to perceive sentiment within the text:

\begin{equation}
v_{\text{fuse}} = v_{\text{word}} \odot v_{\text{affect}}.
\end{equation}

The classifier, which utilizes the features extracted by the feature extractor, distinguishes between real and fake samples. It is a multilayer perceptron consisting of a fully connected layer and a softmax layer.

During the pre-training phase, the feature extractor and classifier are jointly trained to learn initial feature representations and classification abilities. In the fine-tuning phase, an active learning strategy is applied to guide the sample selection process.

Based on the classifier's current predictions, we select samples near the decision boundary, i.e., those with the highest predictive uncertainty. In classification tasks, entropy is adopted as the uncertainty measure for the model’s prediction on each sample. \citet{Sun2025Selecting}

For a binary classification task, given a sample \( x \), the classifier \( D(x) \) outputs the posterior probability of belonging to class 1 as \( P(y = 1 \mid x) \), and to class 0 as \( P(y = 0 \mid x) = 1 - P(y = 1 \mid x) \). The predictive entropy \( \mathcal{H}(x) \) is computed as:

\begin{equation}
\mathcal{H}(x) = -P(y = 1 \mid x) \log(P(y = 1 \mid x)) - P(y = 0 \mid x) \log(P(y = 0 \mid x)).
\end{equation}

Using a threshold \( \tau \), we define the boundary sample set as:
\begin{equation}
\mathcal{B}_t = \{ x \in DU_t \mid \mathcal{H}(x) \geq \tau \}.
\end{equation}

These samples are highly valuable for enhancing model performance. The selected samples are annotated either manually or via automated labeling tools to ensure label accuracy. Once labeled, the samples are forwarded to the data augmentation module, where augmented texts are generated via a hybrid strategy of hard and soft prompts. These augmented samples retain the semantics and labels of the original inputs, thereby increasing the diversity of the training set. The augmented and original labeled samples are jointly used to fine-tune the model in the current training round.

This process is performed iteratively: after each round of fine-tuning, the active learning module selects a new batch of samples for augmentation and training, thereby progressively improving model performance.

\begin{figure}[h!]
    \centering
    \includegraphics[width=0.8\textwidth]{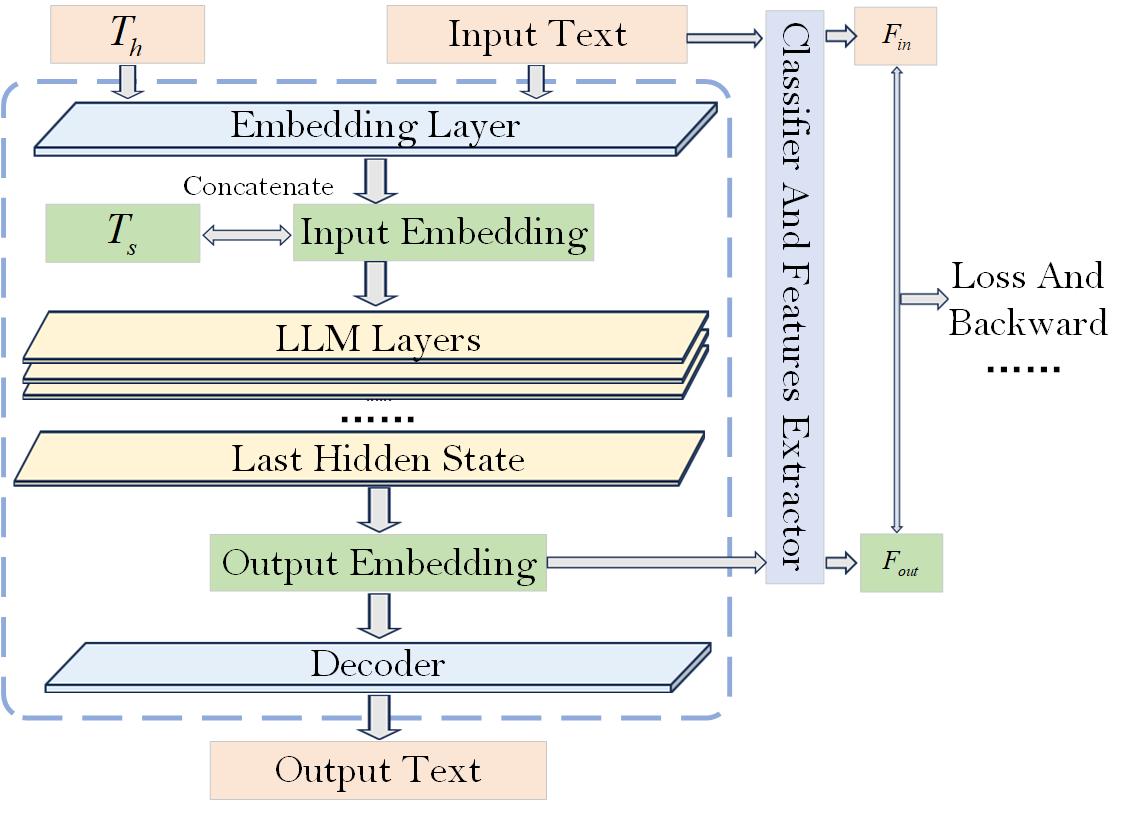}
    \caption{Architecture of Data Augmentation Module}
    \label{fig:aug_module_arch}
\end{figure}

\subsection{Data Augmentation Module}
Figure~\ref{fig:aug_module_arch} illustrates the architecture of the data augmentation module, whose core lies in the training of soft prompts. By learning soft prompts, implicit knowledge from the fake news domain, such as high-frequency argumentation structures, is injected into the Transformer attention mechanism. This effectively guides the large model to perform data augmentation on news samples from specific domains, thereby enhancing the domain adaptability of the large model and achieving fine-grained generation control.

Specifically, we first define a list of hard prompts containing various instructions to guide the large model in generating texts with different styles. These hard prompts are combined with the input text, and after passing through the embedding layer of the large model, we obtain the embedded representation of the input text. Next, we initialize a soft prompt vector and concatenate it with the input text's embedding vector, using the concatenated result as input to the large model. The soft prompt vector, serving as prefix information of the input text, after training, will contain guiding information specific to the task context within the large model.

\begin{equation}
T_{\text{concat}} = T_{\text{in}} \oplus T_h,
\end{equation}
\begin{equation}
E_{\text{combined}} = \text{Embed}(T_{\text{concat}}) \oplus T_s.
\end{equation}

Since the model needs to adapt to news samples from different domains, training soft prompt vectors to effectively guide the large model in performing data augmentation tasks on specific datasets is crucial. Experimental results validate the effectiveness of this method, demonstrating that soft prompt vectors can significantly enhance the model's adaptability and generalization across multiple datasets, thereby improving performance in cross-domain fake news detection tasks.

To avoid the gradient truncation issue that may arise from the softmax and sampling operations during the decoding stage of the large model, we use the vector from the last hidden layer output by the large model \( H_{\text{out}} \) as the embedded representation of the output text. This representation, along with the input text embedding, is fed into the feature extractor to obtain the respective feature vectors. Here, \( F \) denotes the feature extractor.

\begin{equation}
H_{\text{out}} = \text{Hidden}(E_{\text{combined}}),
\end{equation}
\begin{equation}
F_{\text{out}} = F(H_{\text{out}}), \quad F_{\text{in}} = F(\text{Embed}(T_{\text{in}})).
\end{equation}

It is important to note that although our feature extractor is based on BERT encoding, the large models used for data augmentation are typically those that perform better in text generation (such as GPT, Llama, and in our case, zephyr-7b-beta)~\citep{tunstall2023zephyr}. Therefore, directly using text embeddings generated by these large models is not suitable for the BERT-based feature extractor.

\begin{figure}[h!]
    \centering
    \includegraphics[width=0.5\textwidth]{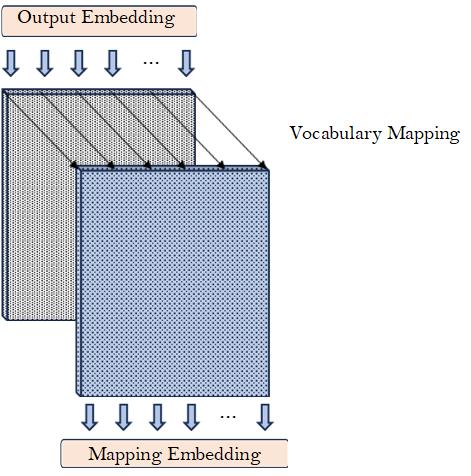}
    \caption{Vocabulary Mapping}
    \label{fig:vocabulary_mapping}
\end{figure}

To address this issue, we first extract the vocabulary of the large model used for data augmentation and establish a mapping relationship with the BERT embedding vocabulary, as illustrated in Figure~\ref{fig:vocabulary_mapping}. Through this mapping, we can convert the text embeddings generated by the large model into BERT-compatible encodings, allowing the generated text to be directly fed into the feature extractor to obtain the text's feature vector.

It is worth noting that the input texts entering the data augmentation module are all located near the decision boundary of the classifier, representing the samples that are most challenging for the classifier to distinguish. The closer the generated augmented samples are to the original input text in the feature space, the more difficult it is for the classifier to classify them, thereby effectively enhancing the model's ability to identify instances near the decision boundary.

Moreover, we explicitly require the large model, through hard prompts, to maintain the main content of the news event unchanged during the generation process, ensuring semantic consistency. This allows the generated augmented text to inherit the label of the input news sample. This design not only avoids the need for additional manual labeling of generated samples but also ensures that the augmented samples extend the decision boundary while maintaining label consistency, thereby further enhancing the model's generalization ability.

Based on this, we first calculate the cosine similarity between the input text and the generated text in the feature space and use it as the first loss objective optimization function. This loss function ensures that the generated augmented samples maintain a high degree of semantic consistency with the input text, while simultaneously enhancing the classifier's discrimination ability near the decision boundary, thereby improving the overall performance of the fake news detection model.

\begin{equation}
\text{Sim}_{\text{cos}}(F_{\text{in}}, F_{\text{out}}) = \frac{F_{\text{in}} \cdot F_{\text{out}}}{\|F_{\text{in}}\| \|F_{\text{out}}\|}.
\end{equation}

To avoid generating trivial samples (i.e., those that entirely overlap with the input in the feature space), we introduce a regularization term. When the Euclidean distance between the input and generated sample falls below a predefined threshold, a penalty is applied, effectively enforcing a minimum distance between them. This promotes diversity in the generated samples and improves the overall quality of augmentation.

Here, \( \lambda_{\text{reg}} \) denotes the regularization weight, \( \delta \) is the minimum acceptable distance, and \( D_{\text{euclid}}(F_{\text{in}}, F_{\text{out}}) \) represents the Euclidean distance:

\begin{equation}
\text{Loss}_1 = -\text{Sim}_{\text{cos}}(F_{\text{in}}, F_{\text{out}}) + \text{Reg},
\end{equation}

\begin{equation}
\text{Reg} = \lambda_{\text{reg}} \cdot \max(0, \delta - D_{\text{euclid}}(F_{\text{in}}, F_{\text{out}})).
\end{equation}

To achieve end-to-end joint training and the collaborative goal of semantic fidelity and decision optimization, we further introduce a second loss function, integrating the training of the data augmentation module and the classifier module. Specifically, the generated text is evaluated by the classifier, and the loss returned by the classifier serves as the second loss function.

\begin{equation}
c = D(H_{\text{out}}), \quad \text{Loss}_2 = (c - c^*)^2.
\end{equation}

Through this design, we ensure that the generated news samples exhibit high uncertainty for the classifier, thereby enhancing the model's ability to learn from hard-to-classify samples. Finally, we perform a weighted summation of the first objective loss (feature space similarity loss) and the second objective loss (classifier loss) to form the total objective loss. We then optimize the soft prompt vectors at the input end of the large model through backpropagation. This optimization process not only improves the generation quality of the data augmentation module but also achieves collaborative optimization with the classifier module, thereby further enhancing the overall performance of the model in fake news detection tasks.

\begin{equation}
\text{Loss}_{\text{total}} = \lambda_1 \cdot \text{Loss}_1 + \lambda_2 \cdot \text{Loss}_2,
\end{equation}

\begin{equation}
T_s \leftarrow T_s - \eta \nabla_{T_s} \text{Loss}_{\text{total}}.
\end{equation}

\subsection{Algorithm Optimization}

The training pipeline consists of two major stages: pre-training and fine-tuning. The model architecture comprises two key modules: a classifier and a data augmentation module. To maximize detection performance, we separately optimize the classifier and the augmentation module during each stage.

\textbf{Pre-training Stage:}  
In the pre-training stage, we begin by training the classifier. The feature extractor encodes both textual and emotional information into feature representations, which serve as inputs to the classifier. The classifier is trained to minimize detection loss, enabling the model to distinguish fake news from real news based on the initial labeled dataset and laying a foundation for subsequent fine-tuning and augmentation.

Following this, the data augmentation module is trained. The goal of this stage is to initialize and optimize the soft prompt vectors, enabling the language model to generate high-quality augmented texts conditioned on the input. Through a dual objective optimization mechanism, the soft prompts enforce semantic consistency between the generated and original texts, while also increasing the sample density near the decision boundary to support the classifier in later stages.

\textbf{Fine-tuning Stage:}

During the fine-tuning phase, we adopt an active learning loop to progressively expand supervision and refine both the prompt-based augmentation module and the classifier. 
At round $t$, the current classifier first scores unlabeled instances in $DU_t$ and selects a batch $S_t$ near its decision boundary for annotation. After obtaining labels for $S_t$, we first fine-tune the soft prompt vector $T_s$ on these newly labeled samples, while freezing all parameters of the LLM backbone. 
This prompt-tuning step adapts $T_s$ to the current data distribution and makes the generation process more task- and domain-aligned.

Next, the augmentation module uses the \emph{fine-tuned} $T_s$ (together with hard prompts $T_h$) to generate an augmented set $S_t'$ from $S_t$, inheriting labels from their corresponding inputs. 
The labeled pool is then updated by merging the newly labeled samples and their augmentations. Finally, we fine-tune the classifier by standard supervised training on the updated labeled pool, so that it can better separate hard instances around the boundary. 
Repeating this procedure for multiple rounds yields a jointly refined augmentation module (via prompt tuning) and classifier (via supervised fine-tuning under the expanded labeled pool), enabling robust detection in label-scarce scenarios.

\begin{algorithm}[H]
\caption{\textbf{Model Pre-training Process}}
\KwIn{Initial labeled pool $DL_0 = (X_L, Y_L)$, Hard prompt $T_h$, Feature extractor $F(\cdot)$, Classifier $D(\cdot)$; Learning rate $\eta$, Number of epochs $N_1$ and $N_2$}
\KwOut{Trained soft prompt vector $T_s$, Updated model parameters of $F$ and $D$}

\textbf{Step 1: Train classifier and feature extractor on $DL_0$}

\For{$epoch = 1$ to $N_1$}{
    Sample $(x_L, y_L) \sim DL_0$;
    \[
    \mathcal{L}_D = \text{Loss}_D\big(D(x_L), y_L\big)
    \]
    \[
    \mathcal{L}_F = \text{Loss}_F\big(F(x_L), y_L\big)
    \]
    \[
    D \leftarrow D - \eta \nabla_D \mathcal{L}_D, \qquad F \leftarrow F - \eta \nabla_F \mathcal{L}_F
    \]
}

\textbf{Step 2: Train soft prompt $T_s$ on $DL_0$}

Initialize soft prompt vector $T_s$;

\For{$epoch = 1$ to $N_2$}{
    Sample $(x_L, y_L) \sim DL_0$;
    \[
    T_{\text{concat}} = x_L \oplus T_h
    \]
    \[
    E_{\text{combined}} = \text{Embed}(T_{\text{concat}}) \oplus T_s
    \]
    \[
    H_{\text{out}} = \text{Hidden}(E_{\text{combined}})
    \]
    \[
    F_{\text{out}} = F(H_{\text{out}}), \quad F_{\text{in}} = F\big(\text{Embed}(x_L)\big)
    \]
    \[
    \text{Sim}_{\text{cos}}(F_{\text{in}}, F_{\text{out}}) = \frac{F_{\text{in}} \cdot F_{\text{out}}}{\|F_{\text{in}}\| \, \|F_{\text{out}}\|}
    \]
    \[
    \text{Loss}_1 = -\text{Sim}_{\text{cos}}(F_{\text{in}}, F_{\text{out}}) + \text{Reg}
    \]
    \[
    T_s \leftarrow T_s - \eta \nabla_{T_s} \text{Loss}_1
    \]
}
\Return Trained soft prompt vector $T_s$, Updated model parameters of $F$ and $D$
\end{algorithm}

\begin{algorithm}[H]
\small
\caption{\textbf{Model Fine-tuning Process}}
\KwIn{Initial labeled pool $DL_0 = (X_L, Y_L)$, Initial unlabeled pool $DU_0 = X_U$, Trained soft prompt vector $T_s$, Hard prompt $T_h$, Feature extractor $F(\cdot)$, Classifier $D(\cdot)$; Learning rate $\eta$, Number of active learning rounds $T$, Query batch size $b$; Number of epochs $N_2$ and $N$ per round; Weights $\lambda_1,\lambda_2$, Target score $c^*$}
\KwOut{Fine-tuned model parameters, Final labeled pool $DL_T$}

Initialize $DL_0 = (X_L, Y_L)$, $DU_0 = X_U$;

\For{$t = 0$ to $T - 1$}{
    \textbf{Step 1: Active learning and labeling}

    Select and query $b$ instances from $DU_t$ according to the active learning strategy, and obtain labels to form
    \[
    S_t = \{(x_u^{(i)}, y_u^{(i)})\}_{i=1}^b.
    \]
    \[
    DU_{t+1} = DU_t \setminus \{x \mid (x,y)\in S_t\}.
    \]

    \textbf{Step 2: Fine-tune soft prompt $T_s$ (loss differs from pre-training)}

    Update $T_s$ by repeating the soft prompt training procedure in Step~2 of Algorithm~1 on the batch $S_t$, except that the objective is replaced by
    \[
    \text{Loss}_{\text{total}} = \lambda_1 \cdot \text{Loss}_1 + \lambda_2 \cdot \text{Loss}_2, \quad \text{Loss}_2 = (c - c^*)^2, \quad c = D(H_{\text{out}}),
    \]
    while all other computations (including $T_{\text{concat}}$, $E_{\text{combined}}$, $H_{\text{out}}$, $F_{\text{in}}$, $F_{\text{out}}$ and $\text{Loss}_1$) follow Algorithm~1;
    \[
    T_s \leftarrow T_s - \eta \nabla_{T_s} \text{Loss}_{\text{total}}
    \]

    \textbf{Step 3: Generate augmented set $S_t'$ from $S_t$ using the updated $T_s$}

    Initialize $S_t' = \varnothing$;

    \For{$i = 1$ to $b$}{
        \[
        T_{\text{concat}}^{(i)} = x_u^{(i)} \oplus T_h
        \]
        \[
        E_{\text{combined}}^{(i)} = \text{Embed}(T_{\text{concat}}^{(i)}) \oplus T_s
        \]
        \[
        H_{\text{out}}^{(i)} = \text{Hidden}(E_{\text{combined}}^{(i)})
        \]
        \[
        T_{\text{out}}^{(i)} = \text{Decode}(H_{\text{out}}^{(i)})
        \]
        Add $(T_{\text{out}}^{(i)}, y_u^{(i)})$ to $S_t'$;
    }

    \[
    S_t' = \{(T_{\text{out}}^{(i)}, y_u^{(i)})\}_{i=1}^b
    \]

    \textbf{Step 4: Update labeled pool and train classifier}

    \[
    DL_{t+1} = DL_t \cup S_t \cup S_t'
    \]

    \For{$epoch = 1$ to $N$}{
        Sample $(x, y) \sim DL_{t+1}$;
        \[
        \mathcal{L}_D = \text{Loss}_D\big(D(x), y\big)
        \]
        \[
        D \leftarrow D - \eta \nabla_D \mathcal{L}_D
        \]
    }
}

\Return Fine-tuned model parameters, Final labeled pool $DL_T$
\end{algorithm}

\clearpage
\section{Experiments}

In the experimental section, we aim to address the following key questions:

\begin{itemize}
    \item Does the SharP model demonstrate significant performance advantages in low resource environments while maintaining satisfactory performance under varying resource conditions?
    \item Does the model exhibit good generalization ability across news data from different domains?
    \item Can the texts generated by the data augmentation module effectively enhance the classifier’s detection performance?
    \item Do different hard prompts and other hyperparameter choices have a significant impact on the model's performance?
\end{itemize}

We systematically evaluate SharP by first describing the experimental setup and then conducting comprehensive tests across various conditions to validate its effectiveness in fake news detection.

\subsection{Experimental Setup}

\textbf{Datasets:}  
In Group A, we evaluate our model on five distinct event-specific datasets: \textit{Germanwings-crash}, \textit{Sydneysiege}, \textit{Ottawashooting}, \textit{Charliehebdo}, and \textit{Ferguson}. These subsets are derived from the PHEME dataset~\citep{zubiaga2016}, which consists of multimodal Twitter data, including users' social information, posted content, and annotated labels. The PHEME dataset is particularly well-suited for fake news detection tasks, as it not only offers rich textual content but also provides sentiment annotations and propagation path information for each tweet. In our experimental design, we proportionally divide the training samples within each subset to comprehensively test model performance under varying sample sizes and across diverse news scenarios.

In addition to the event-specific evaluation on PHEME, we further conduct a cross-benchmark study (Group B) on two widely used misinformation benchmarks, LIAR~\citep{Wang2017LIAR} and Twitter15/16~\citep{Ma2017Twitter15,Ma2017Twitter16}, to assess the robustness of SharP across datasets with different data characteristics. LIAR consists of short, fact-checked political claims annotated with fine-grained veracity labels, which are commonly consolidated into binary labels for fake news detection. Twitter15/16 are event-centric rumor detection benchmarks built from Twitter conversations with reply structures, providing complementary evaluation beyond the PHEME breaking-news threads.

\begin{table}[H]
\centering
\begin{tabular}{lccccc c}
\toprule
Dataset                & Cha. & Fer. & Ger. & Ott. & Syd. & Total \\
\midrule
Labeled Fake News      & 318 & 144 & 95 & 330 & 382 & 1269  \\
Labeled Real News      & 1481 & 719 & 94 & 280 & 559 & 3133 \\
Unlabeled Fake News    & 140   & 140   & 143   & 140   & 140   & 703  \\
Unlabeled Real News    & 140 & 140   & 137   & 140   & 140   & 697  \\
\bottomrule
\end{tabular}
\caption{Dataset statistics for Group A (PHEME) used in our labeled/unlabeled pool setting}
\label{tab:groupA_stats}
\end{table}

\begin{table}[H]
\centering
\begin{tabular}{lcc}
\toprule
Dataset                & LIAR   & Twitter15/16 \\
\midrule
Labeled Fake News      & 5,514  & 445          \\
Labeled Real News      & 6,997  & 429          \\
Unlabeled Fake News    & 143    & 134          \\
Unlabeled Real News    & 137    & 146          \\
\midrule
Total                  & 12,791 & 1,154        \\
\bottomrule
\end{tabular}
\caption{Dataset statistics for Group B (LIAR and Twitter15/16) used in our labeled/unlabeled pool setting}
\label{tab:groupB_stats}
\end{table}

\textbf{Experimental Details:}
To comprehensively evaluate the model's performance under varying resource conditions, we divide the experimental dataset into three subsets: the first group is the labeled data pool used for pre-training, the second group is a fixed-size unlabeled data pool, and the third group is the test data pool.

During the pre-training phase, the model is trained on the labeled data pool to acquire fundamental text classification capabilities. Subsequently, the model proceeds to the fine-tuning phase, where it undergoes further optimization using the unlabeled data pool. In this phase, an active learning strategy is employed to select samples near the decision boundary for labeling and training, thereby enhancing the model's performance through targeted sample selection.

In the experiment, we operationalize \emph{low resource} fake news detection by constraining the size of the labeled training pool. Concretely, for each dataset, we construct a series of label budgets by randomly sampling \{1\%, 5\%, 10\%, 15\%, 20\%, 25\%, 30\%, 35\%, 40\%, 45\%, 50\%, 70\%\} of instances from the labeled pool for training. To reduce sampling bias under small budgets, we adopt stratified sampling to preserve the fake/real ratio (and keep the split as balanced as possible). We then evaluate performance across these budget levels to assess robustness under varying degrees of supervision scarcity, with particular attention to whether the proposed augmentation strategy yields consistent gains when labeled data are limited. Unless otherwise specified, all other settings are kept unchanged across budget levels.

\subsection{Baseline}

In the baseline experiments, we select a series of classical and advanced fake news detection methods as baseline models. We evaluate and compare their performance with the proposed method under the same datasets and experimental settings, with a primary focus on assessing the model robustness and generalization ability in low resource environments.

Here are the baseline models we selected:

\textbf{HAN (2016)}:
A fake news detection method based on the Hierarchical Attention Network (HAN). This method utilizes a hierarchical attention mechanism to model different levels of information in news text (such as words, sentences, and paragraphs), thereby capturing essential semantic features. HAN effectively focuses on critical information within the text and combines structural features of long texts, enhancing the model's classification performance in fake news detection tasks~\citep{yang2016}.

\textbf{dEFEND (2019)}:
A multimodal fake news detection model that integrates textual content and user comments. dEFEND uses explainable attention mechanisms to identify suspicious words and responses, enhancing model interpretability and robustness~\citep{shu2019}.

\textbf{HMCAN (2021)}:
A fake news detection method based on a hierarchical multimodal contextual attention network, which integrates text and social network comment information. It leverages CNNs and hierarchical attention to extract features from multimodal data, aiming to improve classification accuracy~\citep{qian2021}.

\textbf{DDT (2025)}:
A fake news detection method based on active learning, which enhances the model's discrimination ability by combining textual features and emotional features. In addition, DDT employs an active learning strategy, dynamically selecting uncertain samples near the decision boundary for manual annotation and fine-tuning during training. This approach maximizes the model's learning efficiency under limited labeled data conditions. \citep{Hu2025DualAspect}

\textbf{DeepSeek (2024)}:
A general-purpose language model based on the Transformer architecture, incorporating various structural optimization techniques, such as RMSNorm, SwiGLU activation function, and RoPE (Rotary Positional Embedding), to improve performance in long-text modeling and reasoning efficiency. The model is pre-trained on large-scale bilingual (Chinese and English) corpora, demonstrating strong cross-linguistic understanding and generalization capabilities. In this study's baseline experiments, we fine-tune DeepSeek-7B-Base using LoRA (Low-Rank Adaptation) on data with different resource proportions to evaluate its performance and adaptability in the fake news detection task~\citep{liu2024v3}.

\subsubsection{Group A: Event-specific evaluation on PHEME}

To evaluate performance under varying levels of labeled data availability, we consider twelve label-budget configurations: 1\%, 5\%, 10\%, 15\%, 20\%, 25\%, 30\%, 35\%, 40\%, 45\%, 50\%, and 70\%. For each configuration, we randomly sample the corresponding proportion of labeled training data and independently train and evaluate each method on the five PHEME event-specific sub-datasets. This setup allows us to examine performance across different label budgets and event domains.

This experimental setup not only reflects the model’s adaptability under extreme few-shot conditions but also allows for measuring the performance trend as data resources increase. This provides empirical evidence for data augmentation strategies in future low resource text classification tasks.

\begin{table}[H]
\centering
\captionsetup{justification=centering}
\setlength{\tabcolsep}{6pt}
\resizebox{\textwidth}{!}{
\begin{tabular}{llccccccccccccc}
\toprule
Dataset & Stra. & 1\% & 5\% & 10\% & 15\% & 20\% & 25\% & 30\% & 35\% & 40\% & 45\% & 50\% & 70\% \\
\midrule
\multirow{6}{*}{Germanwings-crash} 
 & HAN & 0.500 & 0.734 & 0.713 & 0.691 & 0.691 & 0.745 & 0.734 & 0.755 & 0.723 & 0.766 & 0.745 & 0.713 \\
 & dEFEND & 0.670 & 0.745 & 0.723 & 0.702 & 0.745 & 0.755 & 0.745 & 0.787 & 0.681 & 0.745 & 0.755 & 0.713 \\
 & HMCAN & 0.250 & 0.272 & 0.652 & 0.757 & 0.766 & 0.712 & 0.610 & 0.727 & 0.651 & 0.750 & 0.768 & 0.813 \\
 &DDT & 0.740 & 0.729 & 0.760 & 0.781 & 0.792 & \textbf{0.823} & 0.833 & 0.854 & 0.865 & 0.854 & 0.885 & 0.927 \\
 & Deepseek-7b & 0.617 & 0.617 & 0.617 & 0.617 & 0.628 & 0.628 & 0.628 & 0.617 & 0.606 & 0.617 & 0.617 & 0.606 \\
  & SharP & \textbf{0.760} & \textbf{0.802} & \textbf{0.792} & \textbf{0.833} & \textbf{0.833} & 0.813 & \textbf{0.854} & \textbf{0.885} & \textbf{0.875} & \textbf{0.875} & \textbf{0.917} & \textbf{0.938} \\

\midrule
\multirow{6}{*}{Sydneysiege} 
 & HAN & 0.500 & 0.638 & 0.681 & 0.638 & 0.670 & 0.713 & 0.691 & 0.723 & 0.660 & 0.670 & 0.702 & 0.745 \\
 & dEFEND & 0.606 & 0.660 & 0.628 & 0.617 & 0.681 & 0.691 & 0.660 & 0.691 & 0.702 & 0.691 & 0.702 & 0.681 \\
 & HMCAN & 0.250 & 0.272 & 0.630 & 0.714 & 0.734 & 0.695 & 0.667 & 0.749 & 0.651 & 0.750 & 0.768 & 0.813 \\
 & DDT & 0.677 & 0.708 & 0.771 & 0.771 & 0.760 & 0.771 & 0.823 & 0.865 & \textbf{0.906} & 0.885 & 0.885 & \textbf{0.906} \\
 & Deepseek-7b & 0.462 & 0.441 & 0.462 & 0.462 & 0.462 & 0.462 & 0.462 & 0.462 & 0.462 & 0.462 & 0.462 & 0.462 \\
  & SharP & \textbf{0.760} & \textbf{0.750} & \textbf{0.823} & \textbf{0.802} & \textbf{0.833} & \textbf{0.844} & \textbf{0.833} & \textbf{0.875} & \textbf{0.906} & \textbf{0.927} & \textbf{0.906} & \textbf{0.906} \\

\midrule
\multirow{6}{*}{Ottawashooting} 
 & HAN & 0.500 & 0.649 & 0.681 & 0.723 & 0.638 & 0.670 & 0.713 & 0.606 & 0.638 & 0.681 & 0.691 & 0.660 \\
 & dEFEND & 0.670 & 0.638 & 0.691 & 0.660 & 0.681 & 0.649 & 0.691 & 0.596 & 0.617 & 0.660 & 0.670 & 0.660 \\
 & HMCAN & 0.250 & 0.272 & 0.630 & 0.714 & \textbf{0.766} & 0.737 & 0.667 & 0.727 & 0.651 & 0.727 & 0.750 & 0.813 \\
 & DDT & 0.583 & 0.688 & 0.646 & 0.698 & 0.740 & 0.792 & 0.802 & 0.833 & 0.771 & 0.896 & 0.885 & 0.906 \\
 & Deepseek-7b & 0.500 & 0.500 & 0.500 & 0.500 & 0.489 & 0.489 & 0.500 & 0.489 & 0.489 & 0.489 & 0.489 & 0.489 \\
  & SharP & \textbf{0.695} & \textbf{0.719} & \textbf{0.750} & \textbf{0.813} & 0.740 & \textbf{0.823} & \textbf{0.854} & \textbf{0.844} & \textbf{0.854} & \textbf{0.917} & \textbf{0.896} & \textbf{0.927} \\

\midrule
\multirow{6}{*}{Charliehebdo} 
 & HAN & 0.500 & 0.702 & 0.734 & 0.691 & 0.734 & 0.702 & 0.713 & 0.691 & 0.734 & 0.723 & 0.755 & 0.755 \\
 & dEFEND & 0.691 & 0.691 & 0.702 & 0.638 & 0.723 & 0.702 & 0.755 & 0.638 & 0.734 & 0.702 & 0.691 & 0.766 \\
 & HMCAN & 0.250 & 0.272 & 0.735 & 0.757 & 0.766 & 0.737 & 0.631 & 0.727 & 0.651 & 0.764 & 0.768 & 0.817 \\
 & DDT & 0.708 & 0.750 & 0.792 & 0.792 & 0.823 & 0.833 & \textbf{0.844} & \textbf{0.885} & 0.865 & \textbf{0.865} & 0.896 & 0.885 \\
 & Deepseek-7b & 0.362 & 0.362 & 0.372 & 0.372 & 0.372 & 0.372 & 0.372 & 0.372 & 0.383 & 0.372 & 0.383 & 0.383 \\
  & SharP & \textbf{0.813} & \textbf{0.792} & \textbf{0.823} & \textbf{0.813} & \textbf{0.833} & \textbf{0.844} & \textbf{0.844} & 0.854 & \textbf{0.906} & \textbf{0.865} & \textbf{0.927} & \textbf{0.917} \\

\midrule
\multirow{6}{*}{Ferguson} 
 & HAN & 0.500 & 0.755 & 0.723 & 0.713 & 0.681 & 0.691 & 0.755 & 0.766 & 0.734 & 0.809 & 0.787 & 0.723 \\
 & dEFEND & 0.638 & 0.723 & 0.702 & 0.691 & 0.723 & 0.734 & 0.713 & 0.755 & 0.691 & 0.723 & 0.734 & 0.702 \\
 & HMCAN & 0.250 & 0.272 & 0.735 & 0.757 & 0.745 & 0.737 & 0.610 & 0.727 & 0.651 & 0.764 & 0.768 & 0.817 \\
 & DDT & 0.708 & 0.729 & 0.750 & 0.781 & 0.813 & 0.813 & 0.771 & 0.854 & 0.844 & 0.865 & 0.875 & 0.927 \\
 & Deepseek-7b & 0.681 & 0.681 & 0.692 & 0.681 & 0.692 & 0.692 & 0.692 & 0.702 & 0.692 & 0.702 & 0.692 & 0.681 \\
  & SharP & \textbf{0.792} & \textbf{0.813} & \textbf{0.813} & \textbf{0.833} & \textbf{0.844} & \textbf{0.844} & \textbf{0.865} & \textbf{0.875} & \textbf{0.875} & \textbf{0.883} & \textbf{0.918} & \textbf{0.948} \\

\bottomrule
\end{tabular}}
\caption{Performance comparison on Group A datasets (Acc.)}
\label{tab:baseline_models}

\end{table}

\begin{figure}[htbp]
    \centering
    \subfigure[PCA Projection]{
        \includegraphics[width=0.45\textwidth]{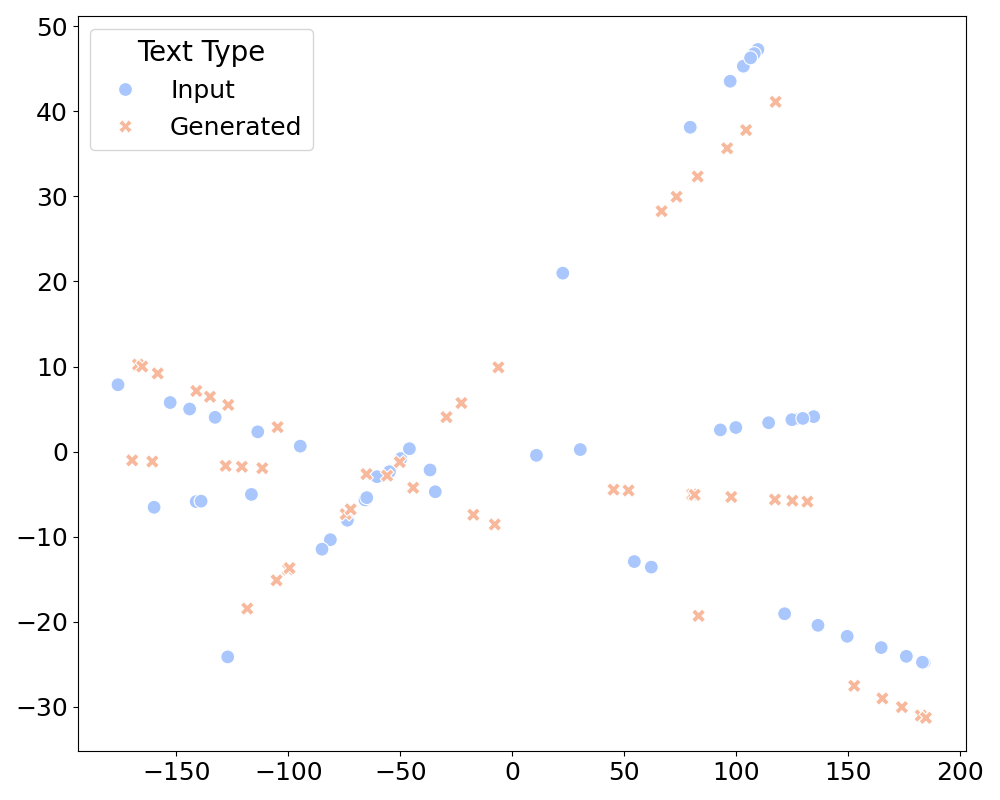}
    }
    \hfill
    \subfigure[UMAP Projection]{
        \includegraphics[width=0.45\textwidth]{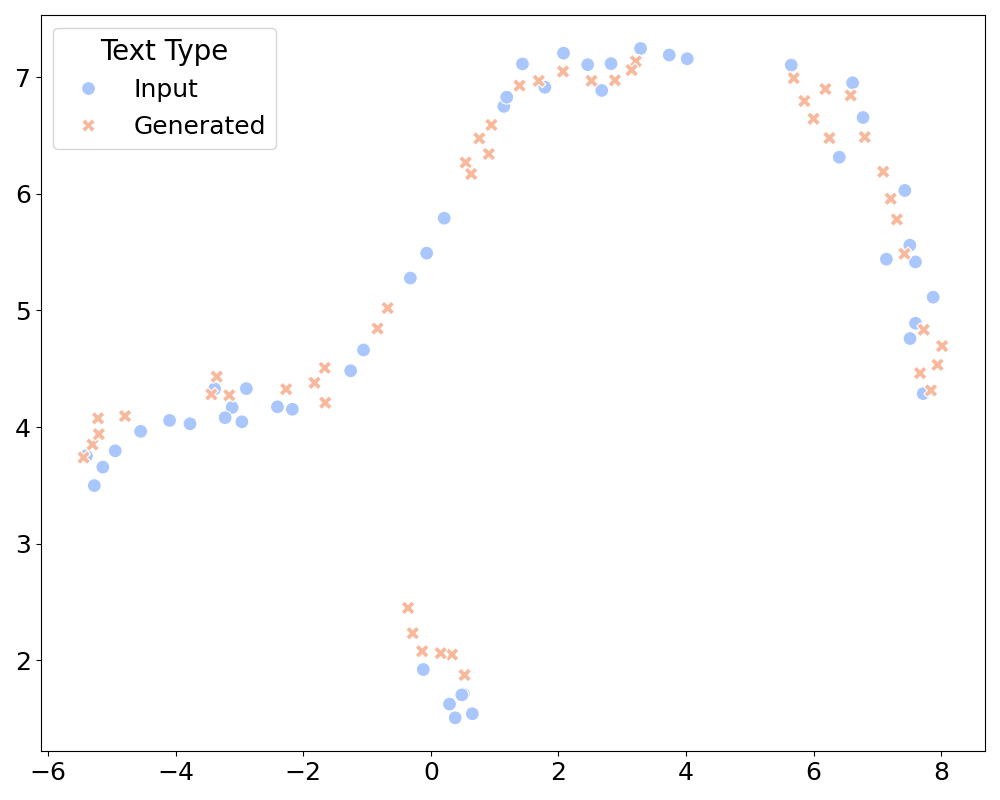}
    }
    \caption{Sample Space Visualization}
    \label{fig:two_images}
\end{figure}

Figure~\ref{fig:two_images} presents the visualization results of the sample space, intuitively showing the distribution relationship between input samples and generated samples in the feature space. The visualization results demonstrate that the trained model is capable of generating enhanced texts that closely align with input samples within the feature space. This verifies the effectiveness of the dual prompt mechanism (hard prompt and soft prompt) and the dual objective optimization strategy (fidelity and discriminability). Furthermore, the results indicate that the generated samples are effectively guided towards the vicinity of the classifier's decision boundary, enhancing the model's ability to learn from hard-to-classify samples, thereby providing the most valuable augmented data for training.

Table~\ref{tab:baseline_models} displays the experimental results of various baseline models. Compared to existing fake news detection methods, the proposed model exhibits significant overall performance improvements, particularly in experimental settings with low sample ratios. The introduction of the data augmentation module effectively improves detection accuracy, fully validating the effectiveness and practicality of the proposed method in low resource environments.

As the number of training samples increases, although the performance gap between models narrows, the proposed method consistently maintains a leading position across multiple datasets and various sample configurations. This result indicates that the proposed augmentation mechanism significantly alleviates the problem of insufficient labeled samples, enhancing the model's robustness and detection capability in data-scarce scenarios.

Moreover, even in cases with sufficient samples, the proposed method still exhibits superior detection performance compared to other baseline models, further demonstrating its universality and advantage in the fake news detection task.

\subsubsection{Group B: Cross-benchmark evaluation }

As a complement to the Group A evaluation, we further assess SharP on LIAR and Twitter15/16 to examine its cross-benchmark generalization and discrimination capability across different misinformation domains and instance formats. 
For LIAR, we collapse the original six-way veracity labels---\textit{pants-fire}, \textit{false}, \textit{barely-true}, \textit{half-true}, \textit{mostly-true}, and \textit{true}---into a binary setting (\textit{Fake} vs.\ \textit{Real}) by mapping \{\textit{pants-fire}, \textit{false}, \textit{barely-true}\} to \textit{Fake} and \{\textit{half-true}, \textit{mostly-true}, \textit{true}\} to \textit{Real}. 
For Twitter15/16, we adopt the commonly used binary rumor verification setting (rumor vs.\ non-rumor) and map it to \textit{Fake}/\textit{Real}, reporting results on the merged Twitter15/16 benchmark. 
We follow the same low resource protocol as in Group A, including identical label budgets (1\%--70\%) with stratified sampling and the same training configuration unless stated otherwise.

\begin{table}[H]
\centering
\captionsetup{justification=centering}
\setlength{\tabcolsep}{6pt}
\resizebox{\textwidth}{!}{
\begin{tabular}{llcccccccccccc}
\toprule
Dataset & Stra. & 1\% & 5\% & 10\% & 15\% & 20\% & 25\% & 30\% & 35\% & 40\% & 45\% & 50\% & 70\% \\
\midrule

\multirow{6}{*}{PHEME (Germanwings-crash)} 
 & HAN         & 0.500 & 0.734 & 0.713 & 0.691 & 0.691 & 0.745 & 0.734 & 0.755 & 0.723 & 0.766 & 0.745 & 0.713 \\
 & dEFEND      & 0.670 & 0.745 & 0.723 & 0.702 & 0.745 & 0.755 & 0.745 & 0.787 & 0.681 & 0.745 & 0.755 & 0.713 \\
 & HMCAN       & 0.250 & 0.272 & 0.652 & 0.757 & 0.766 & 0.712 & 0.610 & 0.727 & 0.651 & 0.750 & 0.768 & 0.813 \\
 & DDT         & 0.740 & 0.729 & 0.760 & 0.781 & 0.792 & \textbf{0.823} & 0.833 & 0.854 & 0.865 & 0.854 & 0.885 & 0.927 \\
 & Deepseek-7b & 0.617 & 0.617 & 0.617 & 0.617 & 0.628 & 0.628 & 0.628 & 0.617 & 0.606 & 0.617 & 0.617 & 0.606 \\
 & SharP       & \textbf{0.760} & \textbf{0.802} & \textbf{0.792} & \textbf{0.833} & \textbf{0.833} & 0.813 & \textbf{0.854} & \textbf{0.885} & \textbf{0.875} & \textbf{0.875} & \textbf{0.917} & \textbf{0.938} \\
\midrule

\multirow{6}{*}{LIAR} 
 & HAN         & 0.318 & 0.436 & 0.564 & 0.564 & 0.439 & 0.564 & 0.406 & 0.564 & 0.482 & 0.557 & 0.466 & 0.564 \\
 & dEFEND      & 0.564 & 0.564 & 0.564 & 0.436 & 0.564 & 0.549 & 0.549 & 0.549 & 0.564 & 0.564 & 0.549 & 0.564 \\
 & HMCAN       & 0.185 & 0.185 & 0.185 & 0.259 & 0.431 & 0.431 & 0.259 & 0.259 & 0.185 & 0.185 & 0.259 & 0.431 \\
 & DDT         & 0.526 & 0.573 & 0.641 & 0.630 & 0.651 & 0.688 & 0.714 & 0.719 & \textbf{0.766} & 0.724 & 0.781 & 0.859 \\
 & Deepseek-7b & 0.583 & 0.582 & 0.585 & 0.585 & 0.588 & 0.586 & 0.584 & 0.587 & 0.585 & 0.584 & 0.578 & 0.565 \\
 & SharP       & \textbf{0.609} & \textbf{0.615} & \textbf{0.662} & \textbf{0.688} & \textbf{0.688} & \textbf{0.703} & \textbf{0.734} & \textbf{0.734} & 0.750 & \textbf{0.740} & \textbf{0.807} & \textbf{0.880} \\
\midrule

\multirow{6}{*}{Twitter15/16} 
 & HAN         & 0.481 & 0.589 & 0.654 & 0.706 & 0.710 & 0.645 & 0.701 & 0.736 & 0.654 & 0.823 & \textbf{0.861} & 0.628 \\
 & dEFEND      & 0.502 & 0.498 & 0.632 & 0.727 & 0.576 & 0.706 & 0.498 & 0.727 & 0.498 & 0.714 & 0.805 & 0.723 \\
 & HMCAN       & 0.250 & 0.250 & 0.333 & 0.250 & 0.500 & 0.333 & 0.667 & 0.500 & 0.500 & 0.667 & 0.750 & 0.750 \\
 & DDT         & 0.687 & 0.748 & 0.786 & 0.809 & 0.765 & 0.791 & 0.826 & \textbf{0.896} & 0.861 & 0.861 & 0.817 & 0.870 \\
 & Deepseek-7b & 0.504 & 0.504 & 0.504 & 0.504 & 0.548 & 0.504 & 0.504 & 0.548 & 0.549 & 0.504 & 0.548 & 0.549 \\
 & SharP       & \textbf{0.783} & \textbf{0.835} & \textbf{0.878} & \textbf{0.844} & \textbf{0.844} & \textbf{0.870} & \textbf{0.844} & 0.870 & \textbf{0.878} & \textbf{0.887} & 0.852 & \textbf{0.896} \\

\bottomrule
\end{tabular}}
\caption{Performance comparison on Group B datasets (Acc.)}
\label{tab:groupB_performance}
\end{table}

As reported in Table~\ref{tab:groupB_performance}, SharP achieves strong performance on both LIAR and Twitter15/16. Across the 24 label-budget configurations evaluated on LIAR and Twitter15/16, SharP achieves the highest accuracy in 21 configurations and ranks second in the remaining three, demonstrating competitive performance across both benchmarks and a range of label budgets. These complementary experiments further demonstrate that SharP generalizes well across heterogeneous benchmarks, including fact-checked political claims from LIAR and conversational microblog threads from Twitter15/16, which differ in domains, topics, and writing styles. This supports the robustness of our approach under diverse benchmarks and input characteristics.

\subsection{Ablation Study}

\begin{figure}[htbp]
    \centering
    \includegraphics[width=0.8\textwidth]{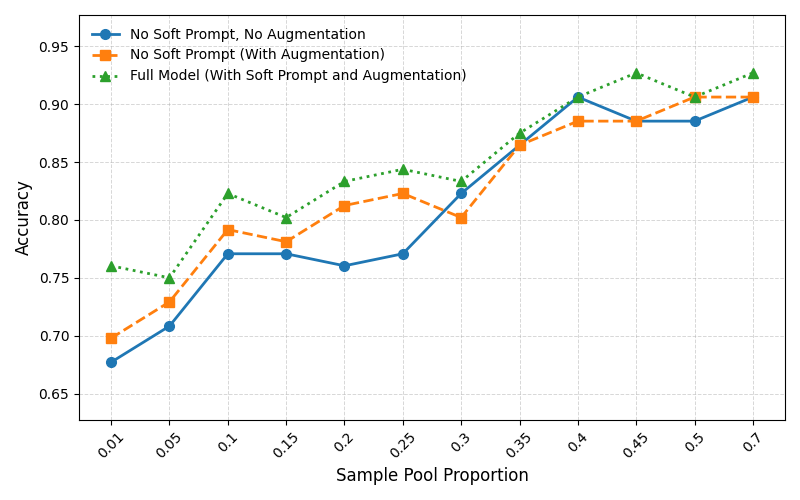}
    \caption{Ablation Study}
    \label{fig:ablation}
\end{figure}

To comprehensively evaluate the model's performance under different component configurations, we designed an ablation study to test the following three scenarios: a single classifier without data augmentation, a large model with data augmentation but without soft prompts, and a large model with data augmentation incorporating the dual prompt architecture. In the experiment, we conducted tests with various proportions of training samples to thoroughly assess the contribution of each component to the model's detection performance.

The results of the ablation study in Figure~\ref{fig:ablation} further confirm the crucial role of the data augmentation module in the model. After incorporating the data augmentation module, the model's performance improved significantly. Furthermore, when the soft prompt guidance was added, the model exhibited further improvements across all evaluation metrics, with the effects being particularly pronounced in low-sample settings. These results demonstrate that introducing large model data augmentation can significantly facilitate model training under low resource conditions. Additionally, the introduction of the dual prompt architecture enables the large model to better understand the generation requirements specific to the current dataset, leading to the creation of higher-quality augmented texts and further enhancing the overall model performance.

\subsection{Hyperparameter Sensitivity Analysis}

\begin{figure}[htbp]
    \centering
    \subfigure[Dual Objective Optimization Weight]{
        \includegraphics[width=0.45\textwidth]{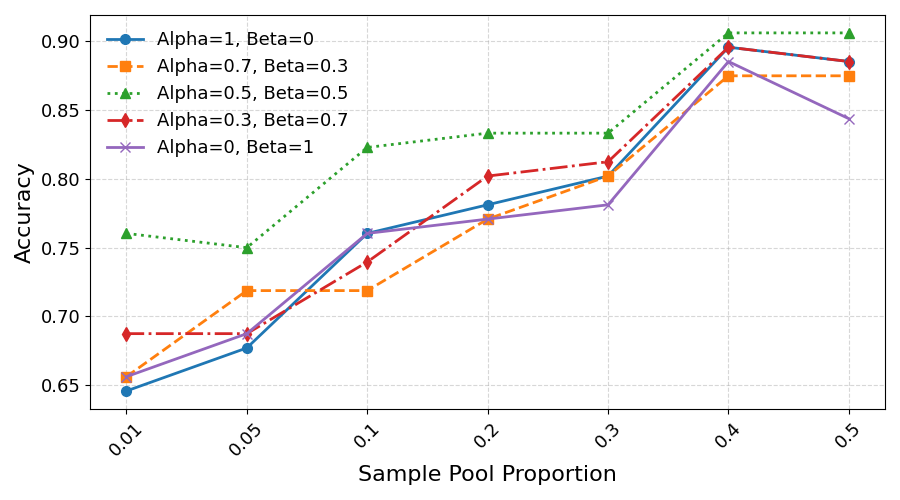}
    }
    \hfill
    \subfigure[Hard Prompt Selection]{
        \includegraphics[width=0.50\textwidth]{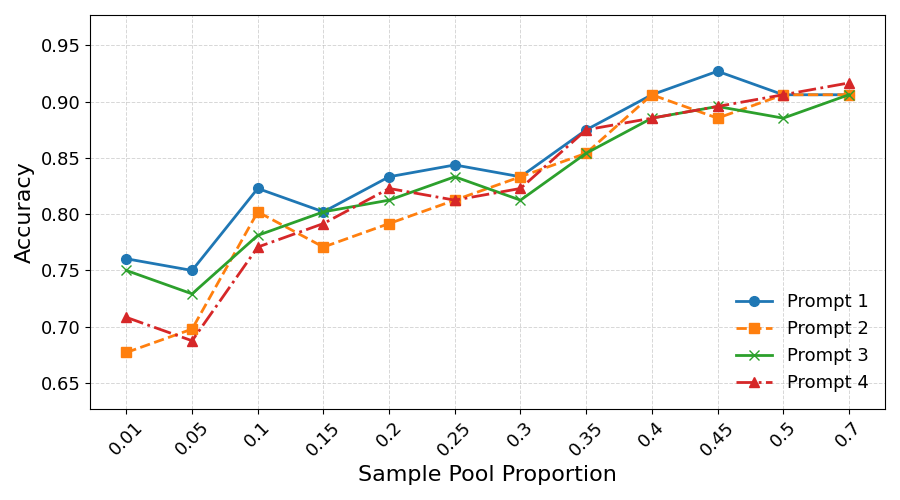}
    }

    \vspace{0.4cm}

    \subfigure[Number of Generated Text]{
        \includegraphics[width=0.46\textwidth]{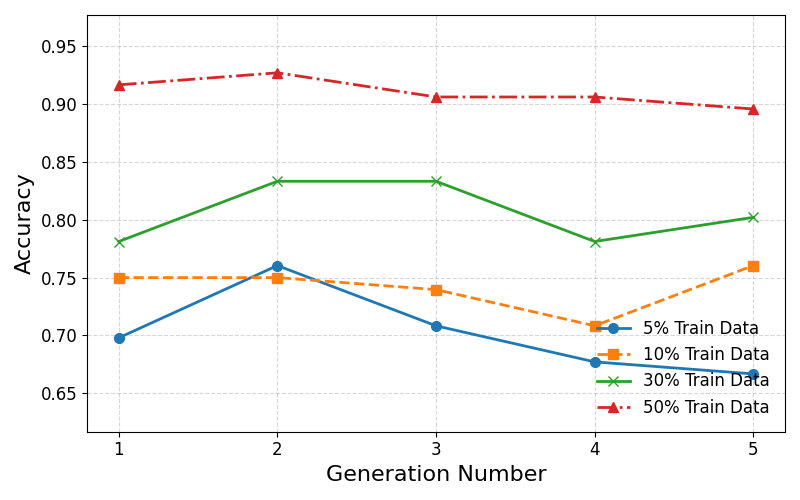}
    }
    \hfill
    \subfigure[Soft Prompt Length]{
        \includegraphics[width=0.46\textwidth]{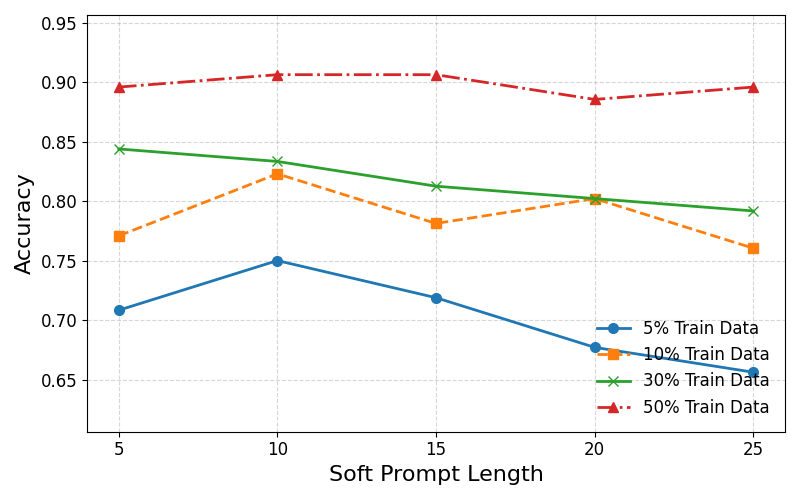}
    }

    \caption{Hyperparameter Sensitivity Analysis}
    \label{fig:four_images}
\end{figure}

To further verify the robustness and stability of our model, we conducted a detailed hyperparameter sensitivity analysis, aiming to evaluate the impact of different hyperparameter settings on model performance. Hyperparameters play a crucial role in training deep learning models as they directly determine the convergence speed, generalization ability, and final detection performance. Therefore, selecting and adjusting hyperparameters appropriately not only enhances the model's performance but also ensures its effectiveness across different tasks and datasets.

In this experiment, we focused on the following key hyperparameters: the choice of hard prompt categories, the size of soft prompt initialization, the number of generated data augmentation texts, and the weight settings of dual objective optimization. Adjusting each hyperparameter can influence the training dynamics of the model, thereby affecting the final detection outcomes.

By systematically analyzing these key hyperparameters, we identified which hyperparameters have the most significant impact on model performance and gained deeper insights into their functional mechanisms. This analysis provides a scientific basis and optimization direction for hyperparameter selection in practical applications. Ultimately, the hyperparameter sensitivity analysis enabled us to identify the most suitable hyperparameter combination for this task, further enhancing the model's overall performance and generalization ability.

Figure~\ref{fig:four_images} presents the experimental results of the hyperparameter analysis. In the analysis of hard prompt category selection, we observed that performance differences between different hard prompt categories were minimal when the number of training samples was large. However, when the number of training samples was small, the performance of certain hard prompt categories was significantly lower.

We hypothesize that this discrepancy may result from the classifier's preference for specific text styles. For example, hard prompt 3 (which uses a more aggressive tone) performed better than hard prompt 4 (which uses a more declarative tone), likely because fake news often employs a more provocative and attention-grabbing style, making such texts easier to identify during classification. As the amount of training data increases, the influence of this preference on model performance gradually diminishes, leading to a reduction in performance differences.

Across the evaluated sample pool proportions, hard prompt 1, which uses simpler and more straightforward expressions, achieved the best or joint-best accuracy in most settings. We therefore adopted hard prompt 1 as the default hard prompt in our experiments.

In the hyperparameter sensitivity analysis, we evaluated the hyperparameters under varying training data proportions (specifically 5\%, 10\%, 30\%, and 50\%) to assess their impact on model performance across different resource levels.

The experimental results indicate that as the initialization length of the soft prompt increases, the model performance gradually improves, reaching a peak before subsequently declining. This phenomenon can be attributed to the following reasons: when the soft prompt length is too short, it fails to adequately encode guiding information for the large model, resulting in lower performance. On the other hand, when the soft prompt length is too long, it occupies a substantial proportion of the input after concatenation with the model's textual input. This excessive additional information hinders the model's ability to effectively understand the task objective, leading to a performance decline.

In the hyperparameter analysis of the number of generated data augmentation texts, we observed that as the generation number increases, the model performance initially rises briefly, reaches a peak, and then continuously declines. Notably, when the training data volume is relatively low, the performance peak occurs with a delay.

This phenomenon can be attributed to the following reasons: when the training data is relatively sufficient, an excessive number of augmented texts can result in a surplus of semantically similar training data in the next training round, thereby inhibiting performance improvement. In contrast, when the training data volume is low, the increase in generated text quantity enhances data diversity. In this case, the addition of more training data primarily contributes to performance improvement, resulting in a delayed arrival of the performance peak.

Finally, in the hyperparameter analysis of dual objective optimization weight settings, we found that models using only a single loss function performed worse under all three sample conditions compared to models that combined both loss functions. When appropriate weights were set and both loss functions were jointly adjusted, the model performance improved significantly. This result demonstrates the effectiveness of dual objective functions in optimizing the quality of generated samples and enhancing the overall model performance.

\section{Conclusion and Future Work}

In this paper, we proposed an LLM-based text data augmentation technique guided by dual prompt engineering, which combines structured hard prompts with learnable soft prompts. This approach was successfully applied to the task of fake news detection under low resource settings. Experimental results demonstrate that the proposed method significantly improves the classifier's discriminative performance by generating diverse samples that are consistent with the domain and preserve the original labels.  Particularly in low resource environments, our framework significantly enhances detection precision and overall classification accuracy by supplying diverse, high-quality augmented training samples.

Beyond empirical performance, our method provides a flexible and model independent solution that can be adapted to various LLM architectures and training setups, making it applicable to a wide range of real world misinformation detection scenarios. In addition, the sample selection strategy, which focuses on uncertain prediction regions, enables more efficient use of annotation resources in active learning settings, further demonstrating the method’s practical value.

In future work, we plan to extend this augmentation framework to a wider range of text classification tasks, including sentiment analysis, topic classification, and intent detection. We are also interested in evaluating the method's effectiveness across multilingual datasets and under distribution shifts. In addition, we aim to explore the integration of multimodal data, such as images and user metadata, to further enhance the model’s contextual understanding and scalability in practical applications like social media content moderation and news verification systems.

\section{Acknowledgements}

This research was supported by the National Natural Science Foundation of China (Grant No. 62176028).

\bibliographystyle{elsarticle-harv}  
\bibliography{references}           

\FloatBarrier
\appendix
\section{Hard Prompt Templates}
\FloatBarrier

\begin{table}[H]
\centering
\caption{Hard prompt template used for style-controlled rewriting, and an example instantiation with the \textit{common} style (the best-performing style token in our experiments)}
\label{tab:hard-prompts}
\footnotesize
\setlength{\tabcolsep}{6pt}
\renewcommand{\arraystretch}{1.10}
\begin{tabularx}{\linewidth}{p{0.23\linewidth} X}
\toprule
\textbf{Item} & \textbf{Content} \\
\midrule
Template $T_h(x,s_t)$ &
{\ttfamily\detokenize{
Rewrite this ONE sentence using <s_t> word choices (minimal substitutions).
Keep meaning/facts; keep names/numbers unchanged.
Return ONE sentence only, no extra text.
<x>
}}\\
Style tokens &
$s_t \in \{\textit{common}, \textit{professional}, \textit{intense}, \textit{neutral}\}$ \\
\specialrule{0.08em}{0.35em}{0.35em} 
\multicolumn{2}{l}{\textbf{Example instantiation (common style)}}\\[-0.2em]
\addlinespace[0.25em]
Prompt ($s_t=\textit{common}$) &
{\ttfamily\detokenize{
Rewrite this ONE sentence using common word choices (minimal substitutions).
Keep meaning/facts; keep names/numbers unchanged.
Return ONE sentence only, no extra text.
The government announced a }\textdollar\detokenize{5 million relief plan on Tuesday.
}}\\
\bottomrule
\end{tabularx}
\end{table}

\end{document}